\documentclass[runningheads]{llncs}
\usepackage[T1]{fontenc}
\usepackage{graphicx}
\usepackage{booktabs}
\usepackage{hyperref}
\usepackage{tabulary}
\usepackage{makecell}
\usepackage[a4paper,margin=2.5cm]{geometry}

\begin{document}
\title{Onto-Epistemological Analysis of AI Explanations}
%
%
\author{
Martina Mattioli\inst{1,2} \and
Eike Petersen\inst{5} \and
Aasa Feragen\inst{5} \and
Marcello Pelillo\inst{1,3,4} \and
Siavash A. Bigdeli\inst{5}
}

\authorrunning{Bigdeli et al.}

\institute{
Ca' Foscari University, Venice, Italy 
\and
Polytechnic University of Turin, Turin, Italy
\and
European Centre for Living Technology, Venice, Italy
\and
Zhejiang Normal University, Jinhua, China
\and
Denmark Technical University, Lyngby, Denmark
}

\maketitle              

\begin{abstract}
Artificial intelligence (AI) is being applied in almost every field.
At the same time, the currently dominant deep learning methods are fundamentally black-box systems that lack explanations for their inferences, significantly limiting their trustworthiness and adoption.
Explainable AI (XAI) methods aim to overcome this challenge by providing explanations of the models' decision process.
Such methods are often proposed and developed by engineers and scientists with a predominantly technical background and incorporate their assumptions about the existence, validity, and explanatory utility of different conceivable explanatory mechanisms.
However, the basic concept of an \emph{explanation} -- what it is, whether we can know it, whether it is absolute or relative -- is far from trivial and has been the subject of deep philosophical debate for millennia.
As we point out here, the assumptions incorporated into different XAI methods are not harmless and have important consequences for the validity and interpretation of AI explanations in different domains.
We investigate ontological and epistemological assumptions in explainability methods when they are applied to AI systems, meaning the assumptions we make about the existence of explanations and our ability to gain knowledge about those explanations.
Our analysis shows how seemingly small technical changes to an XAI method may correspond to important differences in the underlying assumptions about explanations.
We furthermore highlight the risks of ignoring the underlying onto-epistemological paradigm when choosing an XAI method for a given application, and we discuss how to select and adapt appropriate XAI methods for different domains of application.

\keywords{Explainable AI, Epistemology, Ontology, Ethics of AI}
\end{abstract}

\section{Introduction}
The growing use of AI has led to a rising concern for the legitimacy of applying AI tools in different applications.
XAI, to some extent, has been developed for justifying the use of AI tools.
However, AI models can be `explained' in many ways~\cite{Dale2009,Rice2020,Pearl2009}, each performing differently within the completeness/understandability trade-offs~\cite{Kulesza2013,Arrieta2020}.
Through investigation of various XAI tools, the reliability~\cite{arun2021assessing,brunet2022implications}, trustworthiness~\cite{ghassemi2021false,duran2021dissecting,rafferty2022explainable,jabbour2023measuring}, and utility~\cite{miller2023explainable} of particular XAI methods has been challenged.
%
%
While these works focus on [in]validating XAI techniques, the goal of our work is to find which of these requirements are considered when building each XAI method.
We specifically ask: \emph{which philosophical assumptions about the nature and existence of AI 'explanations' are underlying the different proposed XAI methods?}

For a better understanding of XAI methods and their legitimacy, we study their fundamental assumptions about explanations, their existence, and their derivation process.
We study influential XAI methods~\cite{linardatos2020explainable,bai2021explainable,ding2022explainability} and investigate their assumptions about explanations to categorize them with respect to general paradigms of justification.
Specifically, we investigate ontological and epistemological assumptions about explanations in XAI papers, meaning their assumptions about the existence of explanations (ontology) and our ability to gain knowledge \emph{about} those explanations (epistemology). 
In other words, the focus is to gain knowledge \emph{of} an explanation and not about its informative part.

To provide a motivating example, the popular Layer-Wise Relevance Propagation (LRP)~\cite{bach2015pixel} method defines the explanation of ``prediction $f(x)$ as a sum of terms of the separate input dimensions~$x_d$''.
This results in a vectorized explanation such that summing its elements approximates the model output.
This is an ontologically \emph{idealist} approach to explanations, as the authors assert what an explanation is.
It differs from, e.g., \emph{pragmatist} or \emph{reliabilist} approaches, where the explanations are justified through their utility and performance with respect to exterior objects and tasks.
In addition, most of the experiments in the original LRP work~\cite{bach2015pixel} are intended to demonstrate the visual quality of the resulting explanations, indicating a \emph{constructivist} epistemology in which good explanations are ranked through conformity with human understanding.
We will further elaborate on the philosophical terminology and our approach in the following sections.

We begin by discussing what is meant by an AI explanation, and how it differs from model diagnosis and trust analysis.
We then elaborate on the philosophical branches of ontology and epistemology to clarify the space of our investigation.
With these analytical tools in place, we then investigate each XAI method, categorize it in terms of its onto-epistemological assumptions, and provide our reasoning for this categorization.
Our onto-epistemological analysis indicates how popular XAI methods rely on a wide variety of different underlying assumptions, or onto-epistemological paradigms.
We conclude by pointing out how our categorization enables researchers within a specific field to find relevant XAI techniques conforming with their paradigm, and not to struggle needlessly with the validation of methods following other paradigms.
For example, medical research within the scientific paradigm would only be compatible with XAI techniques that are developed or justified within this paradigm, or that are at least compatible with it.
These include methods that believe in the existence of reality exterior to the human mind (i.e., contemporary realist) and are therefore incompatible with mind-dependent paradigms (i.e., interpretivist).


\begin{figure*}[t]
\centering
\includegraphics[width=.98\textwidth]{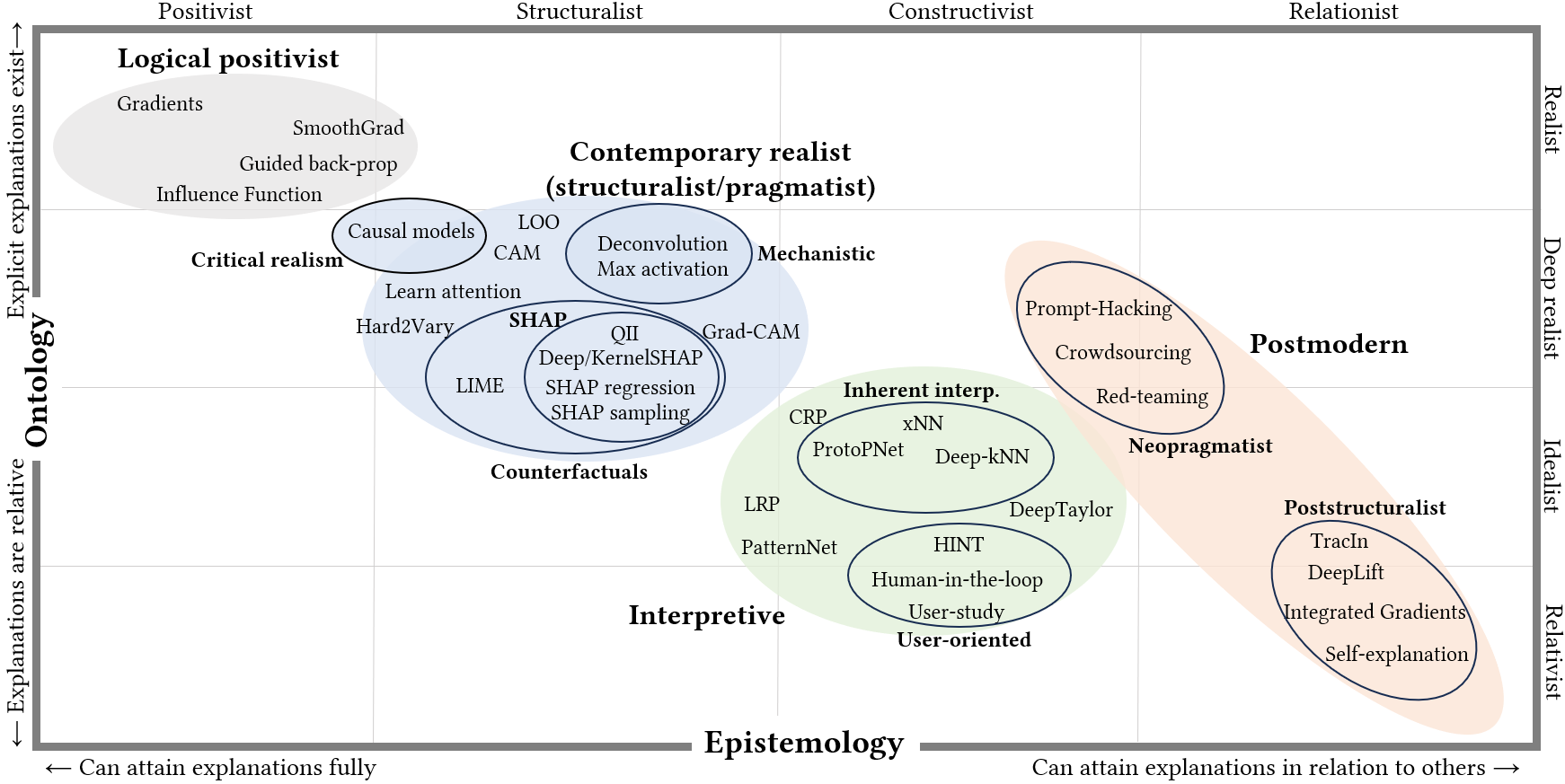}
\caption{Onto-epistemology space of XAI methods.}
\label{fig:ontoepistem}
\end{figure*}

\subsection{Approach and Scope}

We are investigating XAI from the perspective of its assumptions. 
Every XAI method, either explicitly or implicitly, makes assumptions about the existence of absolute explanations, whether or not true and that correct propositions exist that could explain the behaviour of the model for a given input.
This investigation is within the ontology of explanations.
Moreover, XAI tools assume how much of the explanations are attainable by us; to what degree we are able to learn about an explanation.
This is an epistemological analysis of explanation.
Figure \ref{fig:ontoepistem} visualizes a summary of this onto-epistemological space, which we use to place and categorize XAI techniques.
Therefore, we are not focusing on AI explanations as a problem, but as a form of attainable knowledge: While explanations could be used to justify or reject an AI approach, we are concerned with the justification of the explanations themselves (as a second-order knowledge).
Our interest is not to investigate the correctness or utility of these tools from the perspective of the AI/ML sciences.
Furthermore, we are not concerned with the real-world objects that data might represent.

In our analysis, we only consider XAI tools in the context of deep learning research, due to their recent impact and utilization.
For this, we investigate the most well-known works and the original published paper associated with each technique.
In Table~\ref{tbl:paradigm_change}, we show the analysis of the ontological and epistemological features of some gradient-based XAI methods.
We note that some of these works might have been validated through later work in different domains and from different perspectives, and therefore, their onto-epistemological positioning might not be limited to our categorization.

\begin{table*}
\def\arraystretch{1.4}
{
    \begin{tabulary}{\textwidth}{lLLLL}
         & \textbf{Gradients}  \cite{simonyan2013deep} & \textbf{Grad-CAM}  \cite{selvaraju2017grad}                & \textbf{LRP} \cite{bach2015pixel}                     & \textbf{Integrated Gradients}  \cite{sundararajan2017axiomatic}                                        \\
    \Xhline{2\arrayrulewidth}
        \textbf{Method}
        & Back-propagating gradients of output w.r.t. input
        & Weighting the gradients with  activation-map importance
        & Normalizing gradients by their contribution in the activation map
        & Sum gradients of interpolated samples between two inputs
       \\
        \textbf{Ontology}
        & Explanation exists
        & Explanation ~exists
        & Idealized; explanation is one that approximates model output
        & Cannot explain a single input; input is explained w.r.t. a reference point
       \\
        \textbf{Epistemology}
        & Available through gradient chain rule 
        & Justified by localization ability of explanations
        & Justified by visual evidence
        & Discoursive justification; articulating pseudo-reasons
       \\
       \hline
    
        \textbf{Paradigm}
        & Logical positivist
        & Contemporary realist
        & Interpretive
        & Postmodern
       \\
    \hline
    \end{tabulary}
}
\vspace{0.5em}
\caption{Example paradigm change through variations of the same technique. All four methods adapt the backpropagation approach (originally used in \textbf{Gradients}  \cite{simonyan2013deep}) and generate explanations in the form of visual saliency maps. }
\label{tbl:paradigm_change}
\end{table*}

\section{Background}

\subsection{XAI Methods}
We give a brief summary of what explanations mean in the AI community and how they relate to our research.
While explanations can be defined in many different ways and are provided for many distinct purposes~\cite{Miller2019,Dwivedi2023}, an AI explanation is generally considered to be some information about what influences the model's decision, providing a hint about the model's decision criteria.
This information may include the evidence in the input datum (at inference time) as well as the data used to train the model (prior knowledge).
Many types of explanations can be interpreted as a change in direction in the input space that leads to a change in the model's decision; such explanations that explain the output in terms of specific features of the given input are sometimes called \emph{factual} or \emph{evidence-based}~\cite{Stepin2021}.
The hope is that this knowledge can then be used for understanding the decisive characteristics of the input datum that influenced the model's decision.
Some of these explanations are \emph{contrastive} in nature~\cite{Miller2019}: why did the model predict $A$ and not $B$?
This group includes methods such as counterfactual explanations and example-based explanations~\cite{Waa2021}.
More broadly, explanatory information can be gathered for a group of inputs (providing \emph{global} explanations), or for an individual input (providing \emph{local} explanations).
%

Two fundamental tensions are inherent in the provision of AI explanations.
Firstly, explanations are almost never unique: the same model's decision for the same input can be `explained' in many different ways~\cite{Dale2009,Rice2020,Pearl2009}. 
This becomes most obvious in the case of contrastive explanations, where arbitrarily many different counterfactual examples could always be provided as explanations~\cite{Pearl2009,Laugel2023}. 
What renders one of these explanations better than the other?
Secondly, there is a trade-off between explanation \emph{completeness} and \emph{understandability}~\cite{Kulesza2013,Arrieta2020}.
While providing the complete model and training dataset specification to the user might constitute a complete explanation of the model's decision process, it would be incomprehensible to a human recipient.
Conversely, a human-understandable (i.e., interpretable) explanation of the same process will necessarily have to be reductive, i.e., only approximately correct.\footnote[1]{Notice that this is separate from the often-claimed-yet-disputed accuracy-interpretability trade-off: while it may be feasible to construct high-accuracy interpretable models~\cite{Rudin2019}, a \emph{complete} explanation of such a model would usually still not be human-understandable.}

\subsection{XAI and Its Relation to Trust, Fairness, and Justification}
Calls for XAI abound in best-practice guidelines and policy documents, due to XAI's assumed central role in enabling `trustworthy', `fair', and `justified' AI~\cite{Colaner2021}.
The relationships between these different concepts are often not straightforward, however.
AI \emph{fairness} and \emph{justice} are complex and multifaceted concepts, and many of their aspects -- inclusion and exclusion, power differentials, problem framing, data sourcing, agency and (re-)assignment of responsibility -- are completely separate from anything XAI can hope to address~\cite{CostanzaChock2020,Broussard2023,Coeckelbergh2020a,Barocas2023}.
While XAI can, in principle, help uncover direct discrimination or reasons for model underperformance on certain groups, it has also been observed that the aforementioned reductiveness of AI explanations presents the risk of `fairwashed' explanations~\cite{Anders2020,Dimanov2020a}.
The risk of fairwashing also immediately suggests that the link between XAI and \emph{trust} is nontrivial~\cite{Kim2023} and not unquestionably beneficial, since explanations can also mislead users.
Positing that trust evaluates a user's belief in the system for their own purposes, XAI could potentially lead to trustworthy AI if ``(1) the user successfully comprehends the true reasoning process of the model, and
(2) the reasoning process of the model matches the user’s priors of agreeable reasoning''~\cite{jacovi2021formalizing}.
Finally, \emph{justification} is most closely linked to explanation, in the sense that XAI is often used to justify both the use of AI -- the argument being that the system is `trustworthy' because it is `explainable' -- and specific model decisions.
Both of these arguments are questionable, as has been pointed out above: explanations can be misleading or simply not informative, and even in the best case, explainability does not imply trustworthiness or the justifiability of AI use~\cite{Lima2022}.
Moreover, explainability is also not a necessary precondition for justifiability: an epistemic tool does not need to be justified as long as its \emph{use} is justified, as pointed out by Wittgenstein~\cite{wittgenstein1969certainty}.
Justification also relates to the verification of knowledge in epistemology, which is how we will employ the term in this work.

\subsection{Philosophical Background: Ontology and Epistemology}
Beyond our onto-epistemological analysis of AI explanations, explainability has been a topic of a broad philosophical discussion, aimed at investigating how science might answer the `why' questions. Explanations, indeed, have a very wide philosophical tradition, and their history can be traced back for centuries. 
For instance, within this debate, philosophers have posited questions regarding the existence of explanations, their nature, and their role in providing understanding. Once (and if) their existence is assumed, explanations are opposed to descriptions and are considered two distinctive goals of science~\cite{Salmon4decades}.

In this section, we provide an overview of the philosophical terminology used in this work and outline its meaning and relevance in the context of AI explanations.
In particular, we categorize XAI methods according to the ontological and epistemological assumptions that they make about explanations, and we identify certain sets of these assumptions with Kuhnian `paradigms'~\cite{kuhn2012structure}.
Methods originating in different onto-epistemological paradigms are based upon fundamentally different assumptions about explanations.
Therefore, inter-paradigm comparisons and validation of methods may be difficult or meaningless.
In addition, methods developed in one paradigm may not be meaningfully applicable to problem domains founded upon another paradigm.
In the following, we provide a very brief introduction to ontological and epistemological schools of thought that will become relevant for our paradigmatic analysis further below.
We refer the reader to the work of Rabetino et al.~\cite{rabetino2021re} for a summary of paradigmatic reviews in other domains.

\paragraph{Ontology} is the study of what is considered to exist.
It is concerned with ``the most general features and relations of the entities which do exist''~\cite{sep-logic-ontology}. More simply, it relates to what is true and real.
Transposed to the XAI context, this refers to the belief in the existence of propositions that explain the model's behaviour.
Considering the XAI work, \emph{realists} and \emph{deep realists} approaches assume that explanations have an independent existence, as at least a part of reality lives outside the human mind.
\emph{Idealist} approaches, on the other hand, accept the existence of explanations only through mental construction and, therefore, reject their existence outside of a human mind.
Further, the \emph{relativistic} approaches reject the explicit existence of explanations, i.e., they form existence only through relations.
We will discuss these ontological paradigms in the XAI context in more detail in the next section.

\paragraph{Epistemology} is the study of knowledge. It seeks to understand the practices that grant the status of knowledge to beliefs, i.e., knowledge justification~\cite{sep-epistemology}.
In the XAI context, this relates to whether and how (much) we are able to learn about the explanations, namely the knowledge \emph{of} an explanation and not its informative part. This shift of interest allows us to grasp the justification for the use of an XAI model within the respective paradigm. Hence, it is possible to interpret explanations in AI through the existing (epistemological) schools of thought.
Indeed, similarly to assumptions about the ontology of explanations, XAI methods vary in their epistemological assumptions.
\emph{Positivists} assume that full comprehension of reality is possible through sensation, and inference and explanations follow this logical process.
\emph{Structuralists} further assume that explanations are only observable from a standpoint, e.g., a scientific field.
Therefore, they assume that one can understand reality, but not in its complete objectivity.
\emph{Constructivists} believe that reality can only be understood through human mental constructions, opinions, and preferences.
A good explanation is, thus, one that has the most conformity with human understanding (i.e., interpretive). 
\emph{Relationists} assume that we cannot understand explanations as they exist, but only in relation to other explanations.
We will discuss these epistemological paradigms in the XAI context in more detail in the next section.

\vspace{1em}Notice that these two dimensions -- ontological and epistemic assumptions -- are \emph{not} identical to the completeness/understandability dimensions alluded to above: one may, for instance, assume the existence of complete but \emph{relative} explanations (ontological assumption), or that explanations are knowable in principle (epistemic assumption) but may still be too complex to understand for humans.

\section{Onto-Epistemological Paradigms of XAI}
\label{sec:paradigms}

We consider four main paradigms to analyze XAI within the onto-epistemological space.
Figure~\ref{fig:ontoepistem} shows a summary of this space, where each paradigm is placed with respect to its assumptions on ontology and epistemology.
As visualized, there is a strong correlation between ontological and epistemological assumptions in the distribution of paradigms.
This is a logical consequence of ontology, that methods that do not assume a certain reality of the explanations would not focus on understanding them (if absolute explanations do not exist, we simply cannot know them).
Similarly, the development of methods is illogical when we assume limitations in understanding (if we cannot know them, we cannot develop a model for them).
\begin{table*}[t]
\def\arraystretch{1.5}
{
    \begin{tabulary}{\textwidth}{LLLLL}
         & \textbf{Logical positivist} & \textbf{Contemporary realist}                 & \textbf{Interpretivist}                     & \textbf{Postmodernist}                                        \\
    \Xhline{2\arrayrulewidth}
    \raisebox{-4\normalbaselineskip}[0pt][0pt]{\rotatebox[origin=b]{90}{\textbf{Problematic}}} & \textbf{Complete description}: The explanation definition is complete. E.g., gradients of output w.r.t. input~\cite{simonyan2013deep}.
        & \textbf{Subjective description}: description is with respect to a narrow or limited view-point. E.g., LIME~\cite{ribeiro2016should} explains a model `` by approximating it locally with an interpretable model''.
                   & \textbf{Human-idealized description}: Explanation defined by inter-subjective expectations. E.g., in HINT~\cite{selvaraju2019taking} explanations must resemble human saliency maps.
            & \textbf{Rejection or ignorance of explanations}: Not considering explicit explanation of individual model decisions. E.g., Integrated Gradients~\cite{sundararajan2017axiomatic} describe their intention as ``to understand the input-output behavior of the deep network''.                                                    \\
    \raisebox{-3\normalbaselineskip}[0pt][0pt]{\rotatebox[origin=b]{90}{\textbf{Methodology}}}       & \vspace{0pt}  \textbf{Directly from description}: The method is free of assumptions and simplifications.                      & \textbf{Subjective development}: The explanation is built to solve a specific task. E.g. marginalized contributions of input features in SHAP~\cite{lundberg2017unified}. 
                   & \textbf{Follow human understanding}: AI explanations mimic humans'. E.g., HINT~\cite{selvaraju2019taking} uses human annotations to regularize the model.
     & \textbf{Based on comparison and differences}: Explanation is built by making a comparison of multiple inputs/models. E.g. DeepLIFT~\cite{shrikumar2017learning} ``explains the difference in output from some 'reference' output''.
                      \\
    \raisebox{-2.5\normalbaselineskip}[0pt][0pt]{\rotatebox[origin=b]{90}{\textbf{Evaluation}}}
           & \vspace{0pt}  \textbf{Independent}: The method does not rely on results for justifying the explanations.                       & \textbf{Pragmatists (optimizing an objective)}: The explanation is tested w.r.t. a task. E.g., Hard-to-Vary~\cite{parascandolo2020learning} approach ranks explanations based on possible variations.
     & \textbf{Ranking based on human data}: The explanation is evaluated by human observations. E.g., LRP~\cite{bach2015pixel} justifies results through visual comparison.
            & \textbf{Neopragmatists (conformity with human intentions)}: The explanation helps in making/changing external properties. E.g., Prompt-Hacking~\cite{prompthacking} leads to finding systematic flaws of model.
     \\
    \hline
    \end{tabulary}
}
\vspace{0.5em}
\caption{Summary of the indicators we used for finding the correspondence between the research papers and paradigms.}
\label{tbl:approach}
\end{table*}
Table~\ref{tbl:approach} shows a summary of philosophical questions we considered to investigate different sections of the research papers and to find their correspondence with respect to each philosophical paradigm.
It is important to note that there is not always a clear boundary between the paradigms, and a method could be originated somewhere in between.
Additionally, methods could be developed in a multi-paradigmatic setup, where different assumptions are taken at different parts of the research.
Therefore, we focus on the main assumptions that the method uses to justify its development of explanations.

\subsection{Logical Positivist}
\label{sec:positivist}
The contemporary positivism, known as logical positivism, refers to the approach where all knowledge justification must either come from established prior knowledge or from experience.
AI is mostly built on the assumptions of this logical positivist paradigm.
AI tools are ``positivist in the sense that they assert a unique and absolute empirically-accessible external reality that is apprehended by the senses and reasoned about by the cognitive processes''~\cite{vernon2007philosophical}.
Therefore, this group assumes the existence of explanations (ontologically realist) and the ability to attain them completely through observation (epistemologically positivist).

In the XAI context, this paradigm considers the detailed aspects of AI development as the sole way of explaining the tools.
Within the making of the AI models, all of the processes for building deep neural networks and their inference clearly exist, are understandable, and relate to the model and its objective (i.e., training loss).
Therefore, reasoning for the model's decision can be done via analysis of the model with respect to its objective.
For example, a classifier's decision for an input can be explained by looking at the gradients of its objective with respect to the input: The gradients, basically, show where the model draws the decision boundary for that input.

Therefore, as logical positivists, this group believes that AI explanations exist and can be fully attained through the investigation of the model and its objective. We list the exemplary approaches of this group below.

Single-pass backpropagation of model gradients has been used to explain model output~\cite{simonyan2013deep}.
This explanation shows the main input direction in which the model bases its decision.
Attainable through a model, training loss, and optimization strategy (e.g., gradient descent), it encompasses the knowledge of the direction that will make the biggest change in the model's decision.

Guided Back-prop~\cite{springenberg2014striving} is building on the gradients and the class activations to visualize pixel contributions. It thresholds the back-propagated gradients to find regions of the input with a positive contribution to the outcome.
This is built on the foundation of the classifier: gradients of the objective with respect to the input.
Note that this method takes guided gradients only to locate positively contributing directions.
Otherwise, it neither rejects the existence of the negative parts nor rejects the ability to attain them.
Therefore, it is not a realist approach.

Smooth-Grad~\cite{smilkov2017smoothgrad} analyzes the change of output with respect to the variation of the input image by adding Gaussian noise to the input images and averaging the gradients.
Note that this perturbation of the input image is not to make counterfactuals, but to make variations of the factual input to smooth the results.
Although this work uses visual comparison to demonstrate performance, its knowledge justification comes from the kernel density estimate of the gradients centered at the input.

\textit{Influence functions~\cite{koh2017understanding}} were introduced to measure the influence of individual training samples on a single decision of the model.
The training objective is approximated through a second-order expansion around the optimal model, which leads to a trackable computation of the rate-of-change between a training sample weight and the model loss at an arbitrary test input.
Although this approach applies heavy approximations for practicality, they still target the true influence each sample makes at forming the model output for a given input.

\subsection{Contemporary Realist (structuralist/pragmatist)}
\label{sec:realist}

This group is similar to the previous one, where reality encompasses the whole existence (e.g., rejection of mind-dependent existence).
The main difference is that in contemporary realism, not all reality is attainable through observations (neo-realistic) and knowledge is influenced by the subjective mind (epistemologically structuralist)~\cite{sep-scientific-realism}.
For example, knowledge built in critical realism might have roots in other sources of foundation and believed premises (ontologically deep-realistic).
Fredrick Jameson summarizes this paradigm as ``a representation as the reproduction, for subjectivity, of an objectivity that lies outside it''~\cite[preface]{lyotard1984postmodern}.
Other approaches in this group include scientific and pragmatistic studies, where reality is studied from, or for, a specific standpoint \cite{oxfordhb_scientific_realism}.
Therefore, in this paradigm, even though there is a belief that AI explanations exist, they can only be studied through a subjective lens (see e.g. Mittelstadt et al.~\cite{mittelstadt2019explaining}).

\textit{CAM~\cite{zhou2016learning}} uses contribution weights of the model's last layer as the saliency maps.
It uses an average pooling layer to find the contribution weights of the last layer's activation.
The saliency maps are then compared to annotated bounding-boxes to regulate the model for the weakly-supervised learning task, based on which they validate their approach.
\textit{Grad-CAM~\cite{selvaraju2017grad}} combines CAM and guided back-prop for increasing the saliency map resolution.
It uses the localization ability (saliency map w.r.t. object bounding box) for evaluation and therefore, is mainly a pragmatic approach.
%
\textit{Hard-to-vary~\cite{parascandolo2020learning}}
is based on the reliablistic assumption that justified explanations are not arbitrary and are hard to vary.
Therefore, this is methodologically pragmatic and realist.

\textit{Leave-One-out} is a technic for valuating training sample influence on individual inference processes.
The simplest approach is to remove a sample from the training set and re-train the model to compare with the performance of the original and the new model.
This would then indicate how much the sample influences the decision (or loss) of the model.
More efficient variants for deep neural networks, e.g.~\cite{lin2022measuring}, remove a combination of samples and approximate this influence (similar to SHAP). 

\subsubsection{Critical Realism}
Through foundationalism, causal models make relational assumptions about reality, with which they build their learning and inference approaches.
These models are then self-explanatory because of the underlying foundational assumptions.
Unfortunately, due to tractability constraints and strong assumptions about reality, these models are significantly less performative than others and therefore less applicable.

\subsubsection{Counterfactuals}
Counterfactual explanations are used for finding changes in the model outcome, given alternative inputs (or model, e.g., in Shapley regression)~\cite{Miller2019}.
These alternatives are made through explanatory requests that are posed with a sense of relevance (subjectivity)~\cite{rosenberg2019philosophy}.
Several techniques adapt this approach depending on the desired questions and explanatory requests.
Even though these techniques find the model change with respect to input change (counterfactual), they are still providing explanations that correspond solely to the original input data and not with respect to the counterfactuals.
Therefore, explanation through counterfactuals is not a post-structural approach, where the explanation is relative to different inputs.

\textit{LIME \cite{ribeiro2016should}} tries to find the reliability of the model prediction with respect to the task at hand. It synthesizes counterfactuals by cutting out segments of the input to find regions that have the most influence on the results.
%
\textit{Shapley regression values~\cite{lipovetsky2001analysis}} builds alternative models that are blind to certain features and compares their performance with the reference model to see how much each feature contributes to the outcome of the model.
This is a structural way of building hierarchical models to control the feature contributions. They are also only interested in the importance of relevant, useful, and known feature attributes.
\textit{Shapley sampling values~\cite{vstrumbelj2014explaining}} is a more practical approach to finding Shapley values, which marginalizes over the feature of interest by sampling it from all training examples and running the original model.
Lunderberg showed many~\textit{SHAP variants~\cite{lundberg2017unified}}
could be converted to SHAP (SHapley Additive exPlanations) following the Shapley constraints. These include DeepSHAP (DeepLIFT + SHAP), KernelSHAP (Linear LIME + SHAP).
Note that DeepLIFT does not assume a specific reference input, but the SHAP extension assumes variations of the input with respect to the training dataset. This means that it is structurally fixed and not subjective.
\textit{Quantitative Input Inference \cite{datta2016algorithmic}} is a measurement of input features' relevance, while considering their correlation in a causal model.

\subsubsection{Mechanistic}
Some methods investigate neural networks by mechanistically opening the model and investigating the role of individual neurons in the inference (forward pass) by projecting their contributions to the input (backward pass).
These models are mostly made for diagnostic reasons, and only a few of these techniques are made for explanations. 
Deconvolution~\cite{zeiler2014visualizing} is the most popular approach in this group and is considered an explanation approach in the most relevant literature that reverses the projection of layers to visualize the activations of interest.
This visualization then reveals the role of the input features in igniting the neuron of interest.
This explanation approach assumes the neuronal connections, influences, and receptive fields to be the only real explanations (naturalistic ontology).
By investigating individual features and neurons at a time, this method aims to gain subjective explanations (structuralistic epistemology).

\subsection{Interpretative}
Within the interpretive paradigm, the major belief is that explanations do exist, but only in the human mind.
This means that explanations are only mental concepts and do not exist outside of the human observer's mind.
As a result of this ontological assumption, the epistemological approach is also limited to knowledge attainable through construction by the human mind; meaning that it is the human observer who \emph{builds} correct explanations.
In other words, interpretive techniques develop XAI where AI model behaviour can be explained, but within subjective language.
Specifically, the construction of explanations is neither causal nor data-driven; it is built to comply with human observer (i.e., user) expectations.

\textit{LRP \cite{bach2015pixel}} proposes pixel-wise explanations that can approximate the model output.
The authors define the explanation of ``prediction $f(x)$ as a sum of terms of the separate input dimensions $x_d$''.
Based on this interpretive assumption, they build a theory for implementing their gradient-based method.
%
\textit{DeepTaylor~\cite{montavon2017explaining}} builds the explanations through a first-order Taylor expansion of the model output, which is centered at a root point.
This root point is not an arbitrary choice and is optimized to be close to the input data of the model.
This design brings this approach into the interpretive paradigm.
\textit{HINT~\cite{selvaraju2019taking}} incorporates a user input saliency map to regularize the attentions of the network and use them as saliency maps to interpret the model.


\subsubsection{Inherent Interpretations}
Some recent XAI model architectures are designed in a way that mimics human interpretations of the inference process, and therefore are inherently interpretive.
There is a high resemblance between this group and causal models (critical realist).
However, causal models use a causal tree to describe the object ontologically.
In contrast, inherent interpretations create models from subjective constructs to describe the task (e.g., discriminative modeling), not the object itself.


\textit{xNN~\cite{vaughan2018explainable}} learns an additive model of input features. Selecting the addition is the key to model interpretability.
Motivated by how human experts make a decision, 
\textit{ProtoPNet~\cite{chen2019looks}} learns a part-space where the similarity between prototypes and input patches is used for interpretable classification.
%
Similarly, \textit{Deep kNN \cite{papernot2018deep}} uses similarity to known training samples for interpreting a classification.
%
\textit{Concept Bottleneck \cite{koh2020concept}} uses a predefined set of concepts as an intermediate latent space and assumes these information to be explanatory for the work of the model.
These concepts are usually from human interpretations of the task.
\textit{Concept Relevance Propagation (CRP) \cite{achtibat2023attribution}} uses heatmaps based on individual concepts as explanations. Because the heatmaps are based on interpretable concepts, this approach can be considered within this paradigm.

\subsection{Postmodern}
Postmodern philosophy supposes that relations form the whole existence.
This is different from the previous paradigms, in which explanations had some existence, exterior or interior to the human mind.
The postmodern group does not consider the real existence of AI explanations, but is concerned with how they are defined and built in relation to others.
In other words, they only consider changes and relations (of either input, model, or the inference process) to be of real existence (relative ontology).
Additionally, these methods attempt to obtain explanations only by drawing relationships and through comparisons (relational epistemology).
Consequently, postmodern approaches usually justify through \emph{functionality} rather than \emph{function} \cite{baudrillard1966systeme}.
Therefore, discoursive justification~\cite{gerken2012discursive} is often used in these works, where argumentation within its context is used to assert the necessity or desirability of the method.

\subsubsection{Poststructuralist}
As a postmodern paradigm, this group is not interested in absolute explanations of the model and its inference, but only in relative explanations.

\textit{Integrated Gradients \cite{sundararajan2017axiomatic}} integrates the gradients for synthetic images (referred to as counterfactuals) for finding ``an attribution of the prediction at input $x$ \emph{relative} to a baseline input $x'$'' (emphasised here).
Additionally, the method justification is mainly based on features of ``sensitivity'' and ``implementation invariance'' and not the explanation itself.
Therefore, by ignoring or rejecting explicit model explanations, the method is aiming for a subjective description of networks; trying ``to understand the input-output behavior of the deep network''.
\textit{DeepLift~\cite{shrikumar2017learning}} explains in input by explaining ``the difference in output from some 'reference' output''.
By changing input regions with a reference input, it explains when the output of the model changes to the reference output.
Therefore, this method does not consider explaining an individual input; it is then ontologically relativist and epistemologically relationist.
Note that DeepLIFT and Integrated Gradients sometimes use a fixed (e.g., zero) vector as their reference input.
We argue that, unless this choice originates from previously discussed sources (model training/inference, model utility, or human interpretation), this approach cannot be associated with other paradigms (positivist, realist, and interpretivist). See Section~\ref{sec:realist} for an example use of DeepLIFT in SHAP that forms a realist approach.
\textit{TracIn \cite{pruthi2020estimating}} estimates the influence of a training sample on the test sample's performance by integrating the change in test-sample-loss during training whenever the training sample is included in the stochastic update.
This method does not measure the absolute influence of the samples but estimates the difference of influence between initialization of the model and its convergence (Lemma 3.1).
This makes the explanation postmodern because it is computed relative to another (also arbitrary) parametrization of the model.
\textit{Self-explanation \cite{huang2023large}} is a different neopragmatist approach that asks the language model to explain its own output.
In other words, the explanation comes as a comparison between two or more outputs of the model, and it rejects (or ignores) the existence of absolute explanations.

\subsubsection{Neopragmatist: Red-teaming, Crowdsourcing, Prompt-Hacking~\cite{prompthacking}}
This group is less interested in the internal processes and justification of the model and is mostly interested in its use and application.
Neopragmatists accept the existence of an explicit relationship between the objective performance and human ideals (ontologically idealist) and attempt to learn this relation (epistemologically relationist).

Interested in the conformity of the model with respect to ideals, norms, and values, neopragmatist XAI approaches investigate model outcome deviations from such goals, implicitly ignoring the model inference process.
Therefore, they are often targeted for an external use (discovery, justification, control, improvement \cite{ding2022explainability}).
For example, neoragmatists could ask ``How can a chatbot reveal private information?''.
This subjective correspondence is different between interpretivism and neopragmatism, where the former targets to mimic the human inference process but the latter aims to comply with the ideal use of the model.
%
%
%
%
%
%
\textit{Red-teaming and Crowdsourcing}
Most of the crowdsourcing and red-teaming work is directed to characterize fairness; in such cases, their analysis is more of a diagnosis when they address specific questions and tests to validate a model.
However, we are concerned in cases where they are interested in explanations and the \emph{how} in the AI inference.
In practice, mass public launches (e.g., the release of Chat-GPT) and red-teaming attacks are employed not only to detect if a model is faulty (e.g., biases, risks, back-doors) but also to find out \emph{how} the model can be exploited or broken~\cite{redteamming,gptpolitics,bugbounty}.
This \emph{how} is exactly to look at the process explaining model functionality.
%
\textit{Do Anything Network (DAN)~\cite{dan}} and \textit{GPTFuzzer~\cite{yu2023gptfuzzer}} are clear cases of this group, where the goal is to explain model behaviour when receiving jailbreak prompts for bypassing the model security.



\section{Discussion}

\subsection{Related Work}
This analysis is mainly motivated by the philosophical investigation in the work of Rabetino et al.~\cite{rabetino2021re}, who performed an onto-epistemological analysis for reviewing the foundations of the research paradigms in strategic management.
We develop a similar analysis in the domain of XAI by looking at the foundational assumptions in existing methods. 
Besides the difference in the topic, a key analytical difference between the work of Rabetino et al.~\cite{rabetino2021re} and ours, is that they used keyword search on related research articles to find and position different approaches, whereas we investigate the XAI tools by directly looking at the ontological, epistemological, and methodological assumptions for positioning each method.

Mittelstadt et al.~\cite{mittelstadt2019explaining} hint at different schools of thought in XAI and people's ``agreement'' on what an explanation is.
They conclude that the ``majority of work in xAI produces simplified approximations of complex decision-making functions'' which are ``like scientific models'' and not paradigmatic.
Our analysis shows, however, that the underlying assumptions for XAI surpass scientific modifications and simplifications and are as significant as paradigm changes.

Dur{\'a}n~\cite{duran2021dissecting} studies the scientific justifiability of XAI and arrives at the conclusion that none of the current approaches are ``scientific explanations''.
Similarly, Ghassemi et al.~\cite{ghassemi2021false} question the applicability of XAI models in health care, pointing to their failure cases.
Our analysis shows that, even though these approaches might be built through technical upgrades, they could still carry the necessary premises to fall inside the scientific paradigm (contemporary realism).
Their usefulness in a specific scientific field, however, is to be determined through the lens of that scientific field.

Some XAI taxonomies have been proposed, most of which focus on a specific aspect, such as the application of AI~\cite{arya2019one,chari2020explanation}, the technology~\cite{arrieta2020explainable,wang2024gradient}, and aid for decision-making \cite{wang2019designing,groza2020towards}.
Compared to them, the proposed analysis gives an abstract and generalized classification of XAI methods, which enhances key characteristics of taxonomies-for-existing-research, mainly its unambiguousness and completeness~\cite{szopinski2019because}.
As a consequence, this philosophical analysis allows us to push boundaries of current knowledge and mark potential contradictions that otherwise remain hidden.
Table~\ref{tbl:paradigm_change} demonstrates an example of such discovery that was otherwise hidden from the technical or utility of the XAI models.
Additionally, because of the increased number of relations in this analysis (see semantic relations in Table~\ref{tbl:approach} such as describes, justifies, build-on, rejects, conforms), the proposed organization of the XAI techniques surpasses the boundaries of taxonomy and classification and should be more considered as a thesauri or ontology~\cite{hjorland2021information}.

\subsection{Limitations}
Building on the onto-epistemological analysis of explanations, our analysis is affected by epistemology’s limitations~\cite{gabriel2020limits}: 1) Epistemological analysis assumes the world remains static; this might not hold in practice, as XAI methods may be validated or justified in other paradigms that they were originally developed in. Therefore, a more fine-grained classification might include temporal aspects of each paradigm to avoid potential over-generalization.
2) Epistemic justification is subject to norms; Our understanding of AI is heavily affected by our values around it.
Therefore, variability of context may change the acceptable way of justifying XAI.
This seems to be less likely to happen due to the dominating global AI narrative, but nevertheless significant with respect to alternative views. A derivative of the onto-epistemological abstraction of XAI is that users have to engage more deeply with it to perform a classification, compared to clear-cut taxonomies based on application or technique. Although this engagement may raise awareness and competence, it might be redundant for use in common-practice development of XAI.


\subsection{Conclusion}
Our onto-epistemological analysis shows that many XAI tools are developed outside of AI's positivist paradigm (see Section~\ref{sec:positivist}).
Small changes in the algorithmic components can lead to major deviations in terms of underlying assumptions about explanations (see Table~\ref{tbl:paradigm_change} for an example).
Indeed, such changes might be incompatible with the use of the XAI tool.

We suggest that questions regarding the utility and performance of XAI tools imply the need for a common ground within the onto-epistemological space of paradigms.
Investigations of explainable AI have led to criticisms concerning its reliability \cite{arun2021assessing,brunet2022implications}, trustworthiness \cite{ghassemi2021false,duran2021dissecting,rafferty2022explainable,jabbour2023measuring}, and utility \cite{miller2023explainable}, based on the assumption that all models are built conforming to the same onto-epistemological paradigm.
Conforming with Kuhnian incommensurability~\cite{kuhn1997structure}, we see that XAI methods are built within different onto-epistemological paradigms.
This indicates that research and development in this regard, including the works to validate XAI, need to consider these assumptions and their compatibility in their evaluation.
For example, DeepLIFT is built by rejecting single-sample-inference explanations (postmodern).
This is not the case in the scientific research paradigm where model inference on a sample is considered to have explanation (see Section~\ref{sec:realist}).
Therefore, scientific use and validation of DeepLIFT would lead to inter-paradigm contradictions~\cite{hoffman1987critical} on the onto-epistemological assumptions of AI explanations.
Similarly, our results indicate that models with incomplete definitions are incompatible with complete (positivistic) explanations.
This limitation can appear for lossless models (e.g., DIP~\cite{ulyanov2018deep}) and models with surrogate/approximate objectives (e.g., DDPM~\cite{ho2020denoising}), because we lack essential information on the desired objective to be explained.

We demonstrate that the onto-epistemological space enables a more accurate and nuanced analysis of XAI methods compared to the trade-offs between completeness/understandability.
In this regard, we see that breaking the XAI assumptions allows the development of tools within other paradigms, such as medicine.
Studies such that attempt to confirm compatibility with their assumptions could place XAI methods within their own paradigm and justify their use.
Similar to the works of Singh et al.~\cite{singh2020quantitative} and Jalaboi et al.~\cite{jalaboi2023dermatological}, one can validate the utility of explanations (e.g., the interpretive method LRP) in a scientific field or evaluate methods of all paradigms in the interpretive paradigm through user studies and validate their correspondence with human expectations.
Following our analysis of the space of assumptions, developers and end-users should be able to locate their XAI method with greater certainty, thereby achieving better conformity with the desired guidelines and assumptions.

\begin{credits}

\subsubsection{\discintname}
The authors declare no conflicts of interest.
\end{credits}
%
%
%
\bibliographystyle{splncs04}
\bibliography{arxiv}

\end{document}